\documentclass[sigconf,screen]{acmart}
\AtBeginDocument{%
  }

\usepackage{amsmath}

\usepackage{amssymb}
\usepackage{algorithmic}
\usepackage{algorithm}
\usepackage{array}
\usepackage{textcomp}
\usepackage{stfloats}
\usepackage{url}
\usepackage{verbatim}
\usepackage{graphicx}
\usepackage{epsfig}

\usepackage{multirow}
\usepackage{booktabs}
\usepackage{caption}
\usepackage{subcaption}
\usepackage{makecell}

\setcopyright{acmlicensed}
\copyrightyear{2025}
\acmYear{2025}
\acmDOI{XXXXXXX.XXXXXXX}
\acmConference[MM '25]{Proceedings of the 33rd ACM International Conference on Multimedia}{October 27--31,
  2025}{Dublin, Ireland}
\acmISBN{978-1-4503-XXXX-X/2018/06}

\acmSubmissionID{7461}

\renewcommand\footnotetextcopyrightpermission[1]{}

\begin{document}

\title{ResGuard: Enhancing Robustness Against Known Original Attacks in Deep Watermarking}


\author{Hanyi Wang}
\affiliation{%
  \institution{Shanghai Jiao Tong University}
  \country{China}
  }
\email{why_820@sjtu.edu.cn}

\author{Han Fang}
\affiliation{%
  \institution{University of Science and Technology of China}
  \country{China}
}
\email{fanghan@ustc.edu.cn}

\author{Yupeng Qiu}
\affiliation{%
  \institution{National University of Singapore}
  \country{Singapore}
}
\email{qiu_yupeng@u.nus.edu}

\author{Shilin Wang}
\affiliation{%
  \institution{Shanghai Jiao Tong University}
  \country{China}
  }
\email{wsl@sjtu.edu.cn}

\author{Ee-Chien Chang}
\affiliation{%
  \institution{National University of Singapore}
  \country{Singapore}
}
\email{changec@comp.nus.edu.sg}

\begin{abstract}
    Deep learning–based image watermarking commonly adopts an “Encoder–Noise Layer–Decoder” (END) architecture to improve robustness against random channel distortions, yet they often overlook intentional manipulations introduced by adversaries with additional knowledge. In this paper, we revisit the paradigm and expose a critical yet underexplored vulnerability: the Known Original Attack (KOA), where an adversary has access to multiple original–watermarked image pairs, enabling various targeted suppression strategies. We show that even a simple residual-based removal approach, that is, estimating an embedding residual from known pairs and subtracting it from unseen watermarked images, can almost completely remove the watermark while preserving visual quality. This vulnerability stems from the insufficient image-dependency of residuals produced by END frameworks, which makes them transferable across images. To address this, we propose ResGuard, a plug-and-play module that enhances KOA robustness by enforcing image-dependent embedding. Its core lies in a residual specificity enhancement loss, which encourages residuals to be tightly coupled with their host images and thus improves image-dependency. Furthermore, an auxiliary KOA noise layer further injects residual-style perturbations during training, allowing the decoder to remain reliable under stronger embedding inconsistencies. Integrated into existing frameworks, ResGuard boosts KOA robustness, improving average watermark extraction accuracy from 59.87\% to 99.81\%.
\end{abstract}




\maketitle

\section{Introduction}

Digital watermarking \cite{van1994digital} is a widely adopted technique for protecting intellectual property and verifying content authenticity. By embedding imperceptible signals into digital images, it enables ownership verification and forensic analysis in applications such as copyright enforcement and source traceability.

\begin{figure}[t!]
  \centering
  \includegraphics[width=\linewidth]{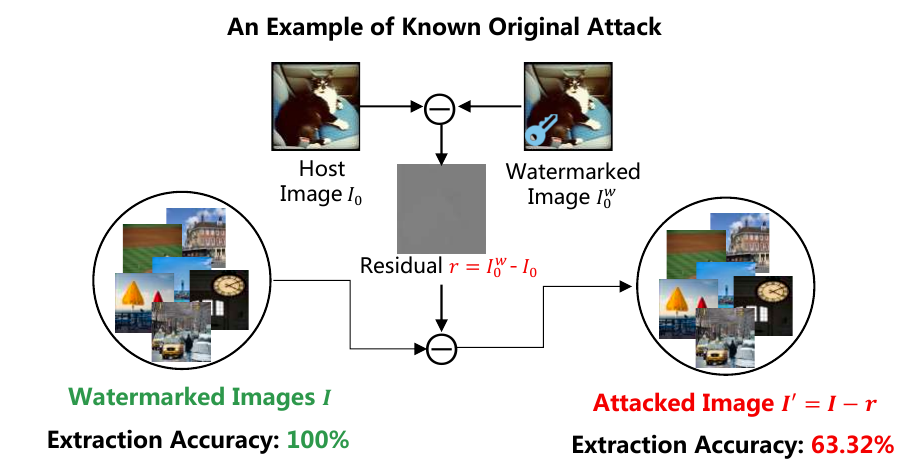}
  \caption{Illustration of an example of the Known Original Attack (KOA) evaluated with RoSteALS~\cite{bui2023rosteals}, showing that even a single residual extracted from one host–watermarked pair can suppress watermark decoding across other images.}
  \label{fig:fig_1}
\end{figure}

A main objective of watermarking is robustness against noise, 
referring to the ability to reliably recover the embedded message under random distortions. To achieve this, deep learning-based watermarking methods~\cite{zhu2018hidden,jia2021mbrs, fang2022pimog} have emerged as an effective paradigm, typically employing an ``Encoder-Noise Layer-Decoder'' (END) architecture. In this framework, an encoder embeds the watermark, the noise layer simulates distortions such as JPEG compression\cite{jia2021mbrs} and screen-to-camera capture\cite{fang2022pimog}, and the decoder recovers the message. Through end-to-end training, these methods achieve substantially higher robustness than traditional coding-based approaches.

Another key objective is robustness against malicious attackers possessing additional knowledge of the decision boundaries. While such scenarios have been widely studied in traditional watermarking, they remain largely underexplored within deep learning frameworks. This gap primarily stems from the difficulty of reformulating analytical, coding-theoretic defenses into differentiable objectives suitable for end-to-end optimization. In this work, we focus on a practical yet challenging adversarial scenario known as the Known Original Attack (KOA)~\cite{cayre2004watermarking,cox2007digital}, where the attacker possesses pairs of original and watermarked images and exploits their differences to remove the embedded watermark.

\vspace{0.1cm}

\noindent 
\textbf{\textit{Known Original Attack (KOA)}} assumes that the adversary has access to a set of host--watermarked image pairs, denoted by $\{(I_i, I_i^w)\}_{i=0}^{N}$, where $I_i$ is the host image and $I_i^w$ is its watermarked counterpart. Leveraging this knowledge, the attacker can estimate a common embedding pattern and remove it from other watermarked images. A simple yet effective method is to compute the average residual
$r_{\mathrm{avg}} = \frac{1}{N}\sum_{i=1}^{N}(I_i^w - I_i)$
and subtract it from a target watermarked image $I^w$ to obtain
$I' = I^w - r_{\mathrm{avg}}$. Empirically, applying this operation can significantly reduce watermark extraction accuracy even when only a single host–watermarked pair is available (Fig.~\ref{fig:fig_1} and Table~\ref{tab:tab_2}).

\vspace{0.1cm}

The success of such attacks reveals a fundamental vulnerability: embedding residuals produced by END frameworks lack strong image-dependency. Instead of forming image-unique patterns tightly coupled to their hosts, residuals remain highly similar across images, making them transferable and easy for adversaries to exploit.

We argue that robust watermarking requires embedding residuals that are highly image-dependent, i.e., inseparable from their host images and difficult to transfer across samples. Building on this insight, we propose ResGuard, a plug-and-play module that injects strong image-dependency into deep watermarking frameworks. ResGuard introduces a residual dependency enhancement (RDE) loss that encourages residuals originating from different host images (but carrying the same message) to diverge, thereby breaking cross-image similarity. To maintain reliable decoding when residual inconsistencies inevitably occur, ResGuard also incorporates a KOA Noise Layer, which introduces residual-style perturbations during training to improve decoder stability under embedding mismatches.

Together, these components substantially reduce residual transferability and close a critical security gap in modern deep watermarking systems.

In summary, we make the following contributions:

\begin{itemize}

    \item We identify a critical yet largely overlooked vulnerability in existing deep learning-based watermarking methods: their susceptibility to the Known Original Attack (KOA).

    \item We propose ResGuard, a plug-and-play module that enhances KOA robustness by enforcing image-specific residual embedding via a novel residual specificity enhancement loss and an auxiliary KOA noise layer.
    
    \item Extensive experiments demonstrate that ResGuard significantly improves robustness against KOA, increasing average watermark extraction accuracy from 59.87\% to 99.81\%, while fully preserving imperceptibility and robustness to common channel distortions.
    
\end{itemize}
\section{Related Work}

\subsection{Traditional watermarking schemes}

Traditional image watermarking typically includes methods based on singular value decomposition (SVD) \cite{mehta2016lwt, soualmi2018schur, su2014blind}, moment-based techniques \cite{hu2014orthogonal, hu1962visual}, and transform domain algorithms \cite{pakdaman2017prediction, hamidi2018hybrid, alotaibi2019text}. Among these, transform domain approaches employing the discrete cosine transform (DCT), discrete wavelet transform (DWT), and discrete Fourier transform (DFT) \cite{kang2003dwt, fang2018screen} are widely adopted. They can embed substantial payloads into images while maintaining invisibility. However, these methods generally demonstrate limited robustness against channel distortions, making the embedded watermarks vulnerable to degradation even under minor alterations to the watermarked images.

\subsection{Deep learning-based watermarking schemes}

Deep learning-based watermarking methods generally offer superior robustness to channel distortions and maintain high visual fidelity. HiDDeN \cite{zhu2018hidden} was the first end-to-end framework to adopt an encoder-decoder architecture combined with a noise layer and an adversarial discriminator. Building on this paradigm, numerous subsequent methods \cite{zhang2019robust, jia2021mbrs, bui2023trustmark, xu2025invismark} have further improved the imperceptibility of watermarks and their resilience to various perturbations. Beyond encoder-decoder approaches, CIN \cite{ma2022towards} and FIN \cite{fang2023flow} introduce flow-based frameworks that leverage invertible neural networks to embed watermarks. Additionally, SSL \cite{fernandez2022watermarking} incorporates watermarks into a self-supervised latent space by relocating image features into a designated region, while RoSteALS \cite{bui2023rosteals} encodes messages directly in the latent space using a frozen VQ-VAE \cite{esser2021taming}.

Despite their robustness to channel distortions, existing deep learning-based watermarking methods remain vulnerable to the Known Original Attack (KOA), where residuals between original and watermarked images can be exploited for watermark removal. To overcome this weakness, we propose ResGuard, a plug-and-play module that enhances KOA robustness while preserving image quality and resistance to common distortions.

\section{Proposed Method}

\begin{figure*}[t!]
  \centering
  \includegraphics[width=\textwidth]{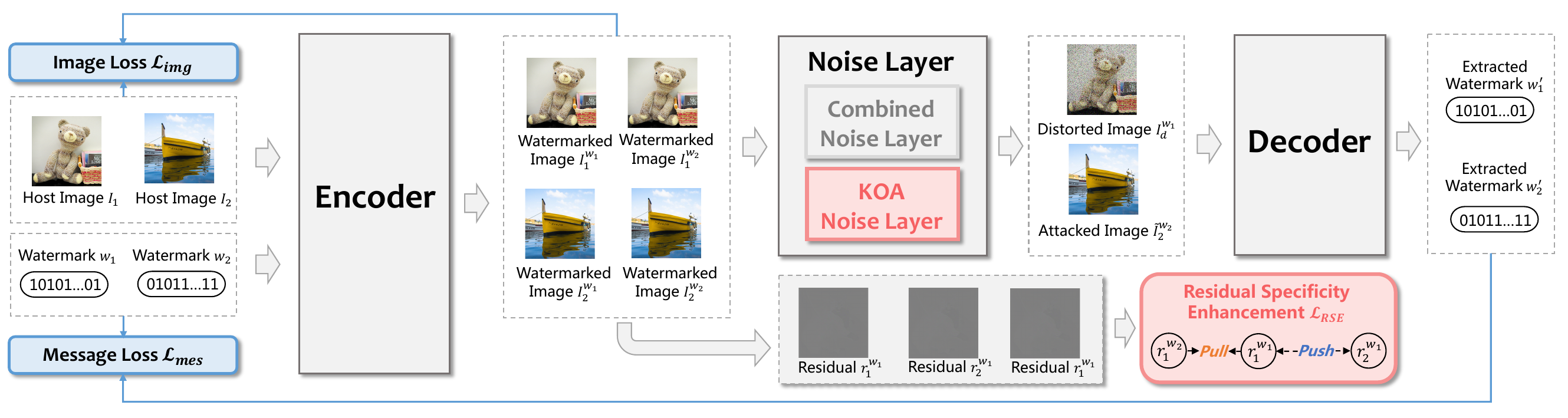}
  \caption{Framework of the proposed ResGuard. It consists of five main components: the encoder $E$, the combined noise layer, the KOA noise layer, the residual specificity enhancement module, and the decoder $D$. The model is trained end-to-end using two basic losses: $\mathcal{L}_{img}$ enforces visual similarity between the host and watermarked images, and $\mathcal{L}_{mes}$ ensures accurate message extraction under both common distortions and KOA attacks. Additionally, $\mathcal{L}_{RSE}$ promotes image-specific residual patterns and suppresses cross-image transferability, enhancing robustness against KOA.}
  \label{fig:fig_2}
\end{figure*}

\subsection{Motivation and Key Ideas}

Our key motivation is to enforce image-specific embedding by ensuring that the residuals of different host images are dissimilar in the pixel domain. In other words, the residuals $(I_1 - I_1^w)$ and $(I_2 - I_2^w)$ should be far apart, preventing a transferable embedding pattern from emerging across images. From another perspective, the goal is to make attacks derived from a known host–watermarked pair non-transferable to other images. More precisely, when an adversary applies the residual extracted from another pair $(I_2, I_2^w)$ to a different image $I$, i.e., forming $I' = I + (I_2 - I_2^w)$, the decoder should still be able to correctly recover the original message $w$.

Guided by these two observations, we develop two complementary training mechanisms. The first corresponds to the former objective and formulates a Residual Specificity Enhancement (RSE) Loss that explicitly encourages inter-image residual dissimilarity. The second introduces a KOA Noise Layer that simulates cross-image residual attacks during training, aiming to minimize their transferability and enhance decoding robustness.

\subsection{Framework Overview}

By integrating the aforementioned solutions, we construct the proposed framework, as illustrated in Fig.~\ref{fig:fig_2}. 
ResGuard employs a dual-pair training strategy in which two host images and two messages are jointly embedded to produce four watermarked images, enabling explicit learning of residual relationships across hosts and messages. 

The framework contains two key components: the Residual Specificity Enhancement (RSE) loss, which explicitly promotes image-dependent embedding to suppress cross-image residual transfer, and the KOA noise layer, which simulates residual-based perturbations to improve robustness against adversarial attacks. Together, these modules train the encoder–decoder pipeline to achieve high watermark extraction accuracy under both natural distortions and KOA conditions.

\noindent
\paragraph{Residual Specificity Enhancement Loss.} To promote image-specific embedding patterns, we aim to ensure that the residuals are primarily determined by the unique content of the host image, rather than by the watermark message. A key strategy is to push apart residuals generated from different host images under the same message, thereby preventing cross-image residual generalization. However, pushing alone is insufficient. Without a consistent reference for image-specific embeddings, the model may lack a stable optimization target and fail to converge toward a well-structured residual space. To address this, we additionally encourage residuals derived from the same host image but carrying different messages to remain close. This contrastive formulation provides a stable anchor for the intrinsic embedding pattern of each image and reinforces the dominance of image content over message content in shaping the residual.

Specifically, given two different host images $I_1$ and $I_2$ and watermark messages $w_1$ and $w_2$, we compute
\begin{equation}
    I_1^{w_1} = E(I_1, w_1), 
    \quad
    I_1^{w_2} = E(I_1, w_2), 
    \quad
    I_2^{w_1} = E(I_2, w_1).
\end{equation}
The corresponding residuals are
\begin{equation}
    r_1^{w_1} = I_1^{w_1} - I_1,
    \quad
    r_1^{w_2} = I_1^{w_2} - I_1,
    \quad
    r_2^{w_1} = I_2^{w_1} - I_2.
\end{equation}
We then define a contrastive loss that pulls together residuals from the same image with different messages and pushes apart residuals from different images with the same watermark message. The loss is given by

\resizebox{\linewidth}{!}{$
    \mathcal{L}_{\text{RSE}} = 
    - \log 
    \frac{
        \exp\left(\text{sim}(r_1^{w_1}, r_1^{w_2}) / \tau \right)
    }{
        \exp\left(\text{sim}(r_1^{w_1}, r_1^{w_2}) / \tau \right)
        + \exp\left(\text{sim}(r_1^{w_1}, r_2^{w_1}) / \tau \right),
    }
$}

\noindent
where $\text{sim}(\cdot,\cdot)$ is the cosine similarity, and $\tau$ is a temperature parameter.

\paragraph{KOA Noise Layer.} To suppress residual transferability, we incorporate a KOA noise layer that explicitly simulates the process of a known original attack during training. This layer applies a residual estimated from one image–watermark pair to a different watermarked image, thereby generating a tampered image. The objective is to encourage the decoder to correctly recover the watermark message even under such residual-based perturbations. 

Specifically, given two distinct host images $I_1$ and $I_2$ embedded with different watermark messages $w_1$ and $w_2$, we obtain
\begin{equation}
    I_1^{w_1} = E(I_1, w_1), 
    \quad
    I_2^{w_2} = E(I_2, w_2).
\end{equation}
We compute the residual for $I_1$ as
\begin{equation}
    r_1^{w_1} = I_1^{w_1} - I_1,
\end{equation}
and apply it to $I_2^{w_2}$ to generate a tampered image
\begin{equation}
    \tilde{I}_2^{w_2} = I_2^{w_2} - r_1^{w_1}.
\end{equation}
Decoding from this tampered image yields
\begin{equation}
    \hat{w}_2 = D(\tilde{I}_2^{w_2}),
\end{equation}
where $D$ denotes the decoder. The KOA loss is then formulated as
\begin{equation}
    \mathcal{L}_{\text{KOA}} = \textit{\text{MSE}}(w_2, \hat{w}_2),
\end{equation}
where $\textit{\text{MSE}}$ denotes the mean squared error.

\subsection{Loss Function}

In addition to our proposed residual-based objectives for enhancing KOA robustness, the standard image loss to ensure invisibility and the message loss to ensure robustness against common distortions are also incorporated.

\paragraph{Image Loss.}
The encoding process embeds the watermark $w_1$ into the host image $I_1$ to produce the watermarked image $I_1^{w_1}$. To preserve visual imperceptibility, the watermarked image is encouraged to remain close to the original host image. This is achieved by minimizing the image loss $\mathcal{L}_{\text{img}}$, defined as
\begin{equation}
\mathcal{L}_{\text{img}}
= \textit{\text{MSE}}(I_1, I_1^{w_1}).
\end{equation}

\noindent
\paragraph{Message Loss.}
The decoding process aims to accurately recover the embedded watermark from potentially distorted watermarked images. To this end, the message loss $\mathcal{L}_{\text{mes}}$ is formulated as
\begin{equation}
\mathcal{L}_{\text{mes}}
= \textit{\text{MSE}}(w_1, \tilde{w}_1) + \lambda_1 \mathcal{L}_{\text{KOA}} ,
\end{equation}

\noindent
where $\tilde{w}_1$ represents the extracted watermark and $\lambda_1$ is a weighting factor that balances the contribution of KOA noise layer during training.

\paragraph{Total Loss.}
The total loss function $\mathcal{L}_{\text{total}}$ is defined as a weighted sum of the message loss $\mathcal{L}_{\text{mes}}$, the image loss $\mathcal{L}_{\text{img}}$, and the residual specificity enhancement loss $\mathcal{L}_{\text{RSE}}$. Formally, this can be written as
\begin{equation}
\mathcal{L}_{\text{total}}
= \mathcal{L}_{\text{mes}}
+ \lambda_2 \mathcal{L}_{\text{img}}
+ \lambda_3 \mathcal{L}_{\text{RSE}},
\end{equation}

\noindent
where $\lambda_2$ and $\lambda_3$are hyperparameters that balance the contributions of each term.

\section{Experimental Results and Analysis}

\subsection{Experimental Settings}

\paragraph{Implementation details.} 
All models are trained on images from the DIV2K \cite{agustsson2017ntire} dataset and evaluated on a test set of 5,000 images from the COCO \cite{lin2014microsoft} dataset. Unless otherwise specified, the combined noise layer applies Gaussian noise by default. The weight factors for the loss function, $\lambda_1$, $\lambda_2$, and $\lambda_3$, are set to 1.0, 0.7, and 0.5, respectively, and the temperature coefficient $\tau$ is set to 0.1. For image size, bit length, and optimization settings, we follow the original training configurations of each watermarking method, which are summarized in Table~\ref{tab:tab_1}. All experiments are conducted using PyTorch 2.5.1 on a single NVIDIA A40 GPU.

\paragraph{Evaluation metrics.} 
To assess the imperceptibility of the watermarked images, we adopt the peak signal-to-noise ratio (PSNR), structural similarity index (SSIM), and learned perceptual image patch similarity (LPIPS). For robustness evaluation, we use bitwise extraction accuracy (Bit ACC.) as the metric.

\begin{table}[t]
\centering
\resizebox{\linewidth}{!}{
\begin{tabular}{llcccc}
\toprule
Methods & Venue & Image size & message length & Optimizer & Learning rate \\
\midrule

HiDDeN  & ECCV 2018 & $128\times128$ & 30  & Adam  & $10^{-3}$ \\
MBRS    & ACM MM 2021 & $256\times256$ & 256 & Adam  & $10^{-3}$ \\
CIN     & ACM MM 2022 & $128\times128$ & 30  & Adam  & $10^{-4}$ \\
RosteALS& CVPRW 2023 & $256\times256$ & 100 & AdamW & $8\times10^{-5}$ \\
InvisMark & WACV 2025 & $256\times256$ & 100 & AdamW & $10^{-4}$ \\

\bottomrule
\end{tabular}
}
\vspace{6pt}
\caption{Training configurations for each baseline method. We follow the original implementation settings to ensure fair comparisons across methods.}
\label{tab:tab_1}
\end{table}
\paragraph{Baselines.} 
We conduct experiments on five representative deep learning-based watermarking methods, including HiDDeN\footnote{\url{https://github.com/ando-khachatryan/HiDDeN}}\cite{zhu2018hidden}, MBRS\footnote{\url{https://github.com/jzyustc/MBRS}}\cite{jia2021mbrs}, CIN\footnote{\url{https://github.com/rmpku/CIN}}\cite{ma2022towards}, RoSteALS\footnote{\url{https://github.com/TuBui/RoSteALS}}\cite{bui2023rosteals}, and InvisMark\footnote{\url{https://github.com/microsoft/InvisMark}}\cite{xu2025invismark}, all of which have publicly available implementations. Our method integrates the proposed ResGuard module into the original architectures of these baseline methods without modifying their network structures. By only altering the training process, we enhance their robustness against KOA.

\subsection{Comparison of the KOA Robustness}

\begin{figure*}[htbp]
    \centering
    \includegraphics[width=0.19\textwidth]{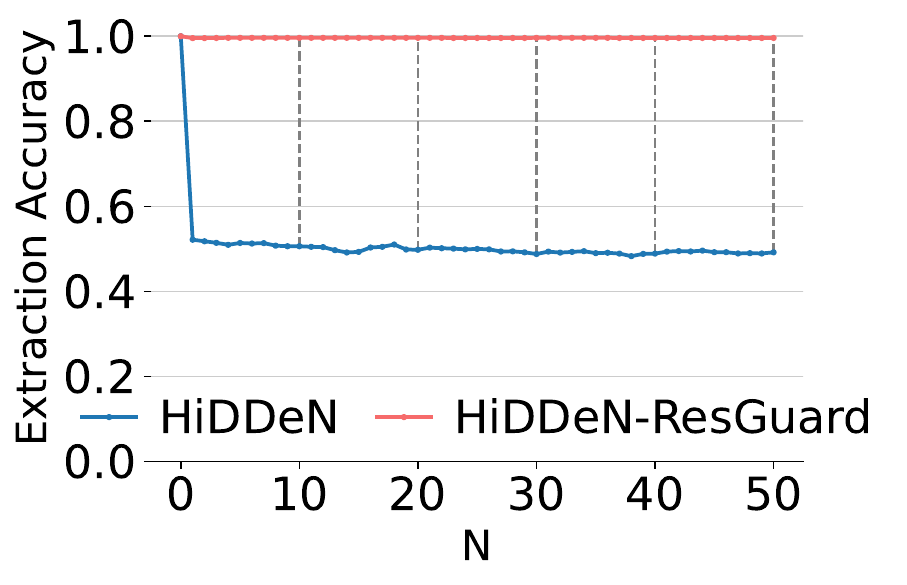}
    \includegraphics[width=0.19\textwidth]{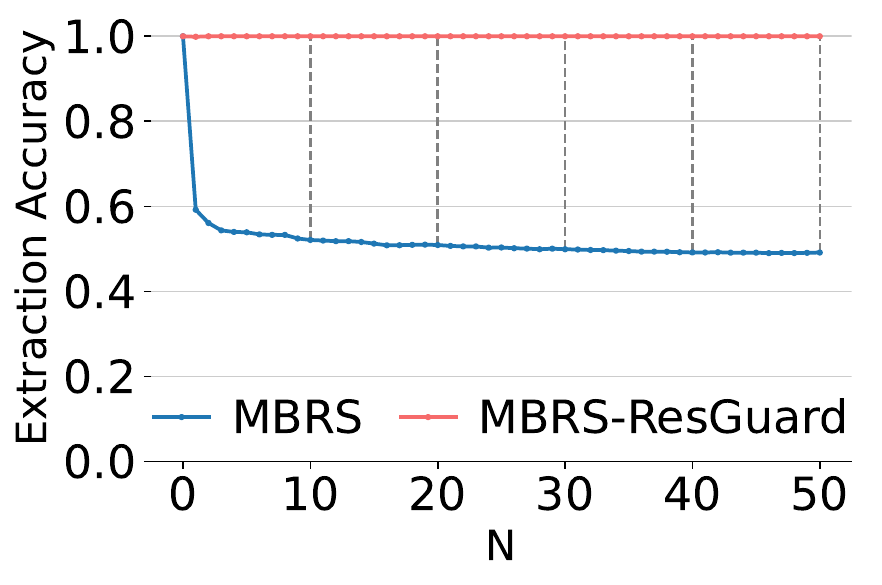}
    \includegraphics[width=0.19\textwidth]{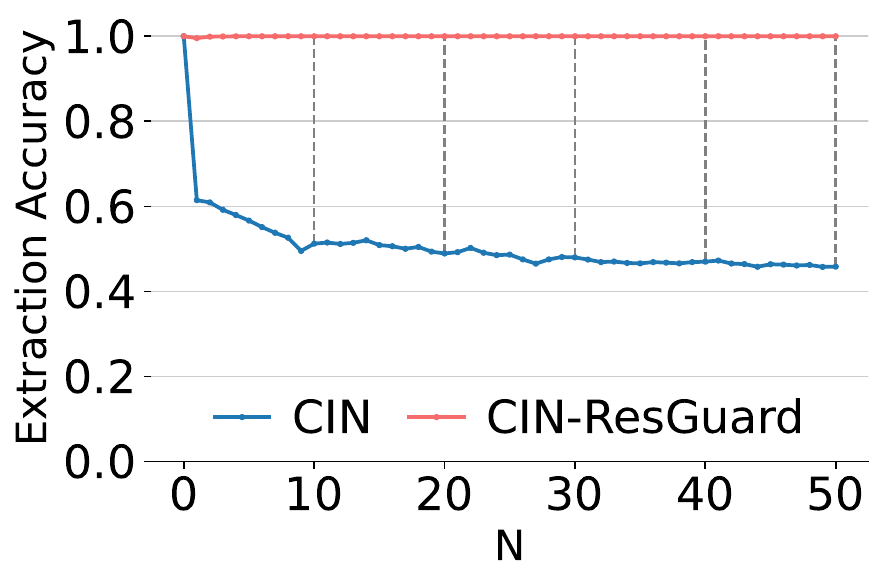}
    \includegraphics[width=0.19\textwidth]{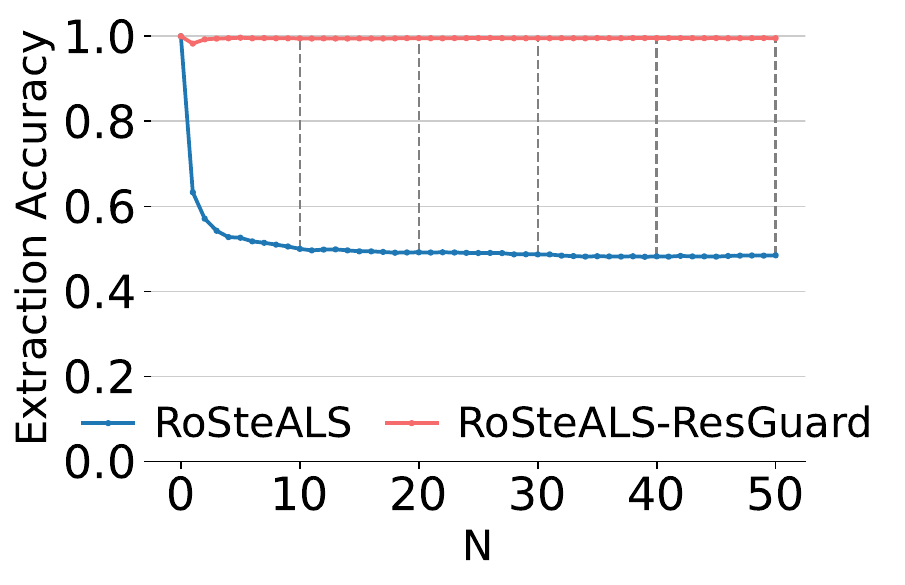}
    \includegraphics[width=0.19\textwidth]{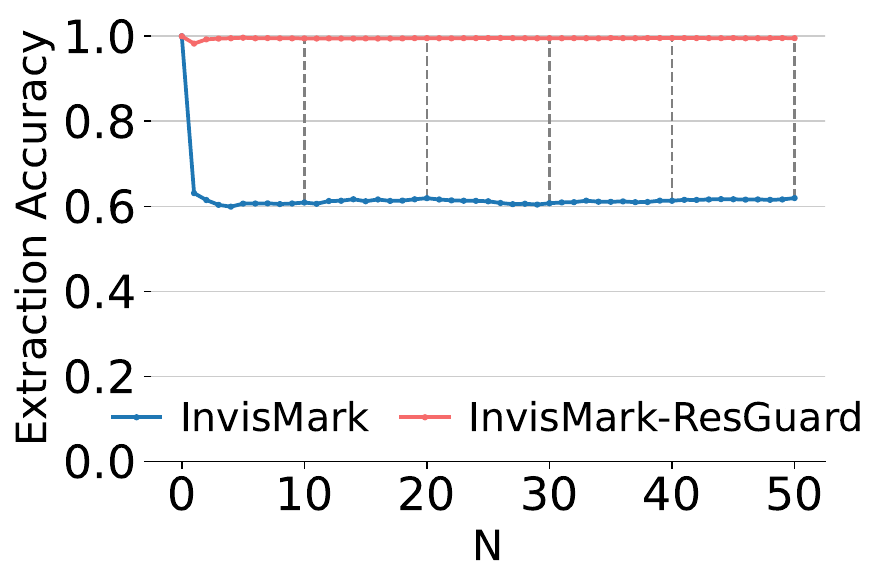}

    \caption{Bitwise extraction accuracy under KOA across five baseline methods. Each curve shows performance variation with the number of available host--watermarked pairs \(N\). Models equipped with ResGuard maintain consistently high accuracy, demonstrating strong robustness against KOA.}
    \label{fig:fig_3}
\end{figure*}

\begin{table}[t]
\centering
\resizebox{\linewidth}{!}{
\begin{tabular}{l|cc|c}
\toprule
 \multirow{2}{*}{Methods} & \multicolumn{2}{c|}{Extraction Accuracy} & \multicolumn{1}{c}{KOA Robustness} \\
 
\cmidrule(lr){2-4} 
 
 & Cln. $\Uparrow$ & Dis. $\Uparrow$ & Bit Acc. $\Uparrow$ \\ 
 
 \midrule
 
 HiDDeN & 1.0000 & 1.0000 & 0.5217 \\
 
 \textbf{HiDDeN-ResGuard} & 1.0000 & 1.0000 & \textbf{0.9957} \\    
 
 \midrule
 
 MBRS & 1.0000 & 1.0000 & 0.5927 \\
 
 \textbf{MBRS-ResGuard} & 1.0000 & 1.0000 & \textbf{0.9988} \\
 
 \midrule
 
 CIN & 1.0000 & 0.9997 & 0.6147 \\
 
 \textbf{CIN-ResGuard} & 1.0000 & 0.9997 & \textbf{0.9983} \\
 
 \midrule
 
 RosteALS & 1.0000 & 0.9926 & 0.6332 \\

 \textbf{RosteALS-ResGuard} & 1.0000 & 0.9931 & \textbf{0.9987} \\
 
 \midrule
 
 InvisMark & 1.0000 & 1.0000 & 0.6312 \\ 
 
 \textbf{InvisMark-ResGuard} & 1.0000 & 1.0000 & \textbf{0.9992} \\

\bottomrule
\end{tabular}
}
\vspace{6pt}
\caption{Comparison results of KOA robustness. “Cln” indicates results on clean images, while “Dis” represents results under channel distortions.}
\label{tab:tab_2}
\end{table}
\begin{table*}[h]
\centering
\resizebox{\linewidth}{!}{
\begin{tabular}{l|ccc|cccccc|c}
\toprule

\multirow{5}{*}{Methods} 
& \multicolumn{9}{c|}{Watermarking Performance} 
& \multicolumn{1}{c}{KOA Robustness} \\

\cmidrule(lr){2-10} \cmidrule(lr){11-11}
& \multicolumn{3}{c|}{Image Quality} 
& \multicolumn{6}{c|}{Extraction Accuracy}  
& \multirow{3}{*}{Bit Acc. $\Uparrow$} \\

\cmidrule(lr){2-4} \cmidrule(lr){5-10}

& \multirow{2}{*}{PSNR $\Uparrow$} & \multirow{2}{*}{SSIM $\Uparrow$} & \multirow{2}{*}{LPIPS $\Downarrow$} 
& \multirow{2}{*}{Cln. $\Uparrow$} & \multicolumn{5}{c|}{Dis. $\Uparrow$} & \\

\cmidrule(lr){6-10}

& & & & & JPEG & GN & S\&P & GB & MB & \\

\midrule

MBRS & 39.62 & 0.9361 & 0.0214 & 1.0000 & 0.9969 & 0.9967 & 0.9941 & 0.9978 & 0.9939 & 0.5713 \\

\textbf{MBRS-ResGuard} & 39.43 & 0.9287 & 0.0297 & 1.0000 & 0.9984 & 0.9988 & 0.9927 & 0.9974 & 0.9983  & \textbf{0.9982} \\

\bottomrule
\end{tabular}
}
\vspace{6pt}
\caption{Evaluation results under diverse channel distortions using MBRS. “Cln” indicates results on clean images without distortions, while 
“JPEG”, “GN”, “S\&P”, “GB”, and “MB” correspond to JPEG compression, Gaussian noise, salt-and-pepper noise, Gaussian blur, and median blur, respectively.}
\label{tab:tab_3}
\end{table*}

We begin by analyzing the robustness of watermarking models against the \textit{Known Original Attack (KOA)}. 
To quantify the effect of the attack strength, we vary the number of available host–watermarked pairs $N$ that the attacker can access, ranging from 1 to 50. For each $N$, the attacker estimates the average embedding residual and subtracts it from the test watermarked images to obtain the attacked results. The bitwise extraction accuracy is then measured to assess decoding reliability under different attack conditions.

As shown in Fig.~\ref{fig:fig_3}, the bit accuracy of baseline watermarking models drops rapidly as the number of available host--watermarked pairs \(N\) increases, revealing that their embedding residuals share strong cross-image similarities that can be easily exploited by the adversary. The accuracy decline gradually saturates when \(N\) becomes large, indicating that once the attacker collects sufficient pairs, the dominant shared residual component is already captured and further samples provide little additional benefit. In contrast, our method consistently maintains high extraction accuracy across all values of \(N\), demonstrating that ResGuard effectively enforces image-specific embedding and suppresses residual transferability. These results confirm that ResGuard substantially mitigates the inherent vulnerability of existing deep watermarking models to KOA by decoupling embedding residuals from cross-image correlations.

In addition, we observe that the bit accuracy already drops significantly when $N=1$, indicating that the attacker can effectively suppress the embedded watermark even with access to a single host–watermarked pair. This finding highlights the severe vulnerability of existing deep watermarking models to the minimal-case scenario of KOA, where the knowledge requirement for the adversary is extremely low. Consequently, in all subsequent experiments, we adopt $N=1$ as the default attack setting to simulate the most realistic and challenging case.

To ensure a fair comparison, all baseline methods are retrained both in their original form and with our ResGuard module under identical training configurations. Since KOA essentially operates by subtracting an additive residual, which can be viewed as introducing a quasi-random perturbation during image transmission, we configure the noise layer for all models to include only Gaussian noise. This setup allows us to examine whether improving robustness to Gaussian noise, as a typical random distortion, translates into resilience against KOA. The results are summarized in Table~\ref{tab:tab_2}.

Experimental results demonstrate that our approach significantly enhances robustness against KOA while fully preserving the watermark extraction accuracy of the baseline methods. 
Across all baselines, our ResGuard-enhanced models achieve substantial improvements in bit accuracy under KOA conditions, attaining an average extraction accuracy of 99.81\%. 
Specifically, we observe relative improvements of 47.83\%, 40.73\%, 38.53\%, 36.55\%, and 36.88\% over HiDDeN, MBRS, CIN, RoSteALS, and InvisMark, respectively. Notably, under standard Gaussian noise, all methods maintain near-perfect accuracy, yet their performance drops markedly to approximately 50\% when subjected to KOA. This discrepancy highlights that robustness to Gaussian noise does not directly imply robustness against KOA, underscoring the necessity of explicitly enforcing image-specific embedding. To this end, our approach is designed as a plug-and-play solution that integrates seamlessly with existing deep learning-based watermarking methods, without requiring modifications to their original architectures or intrinsic properties.

\begin{figure}[t!]
  \centering
  \includegraphics[width=\linewidth]{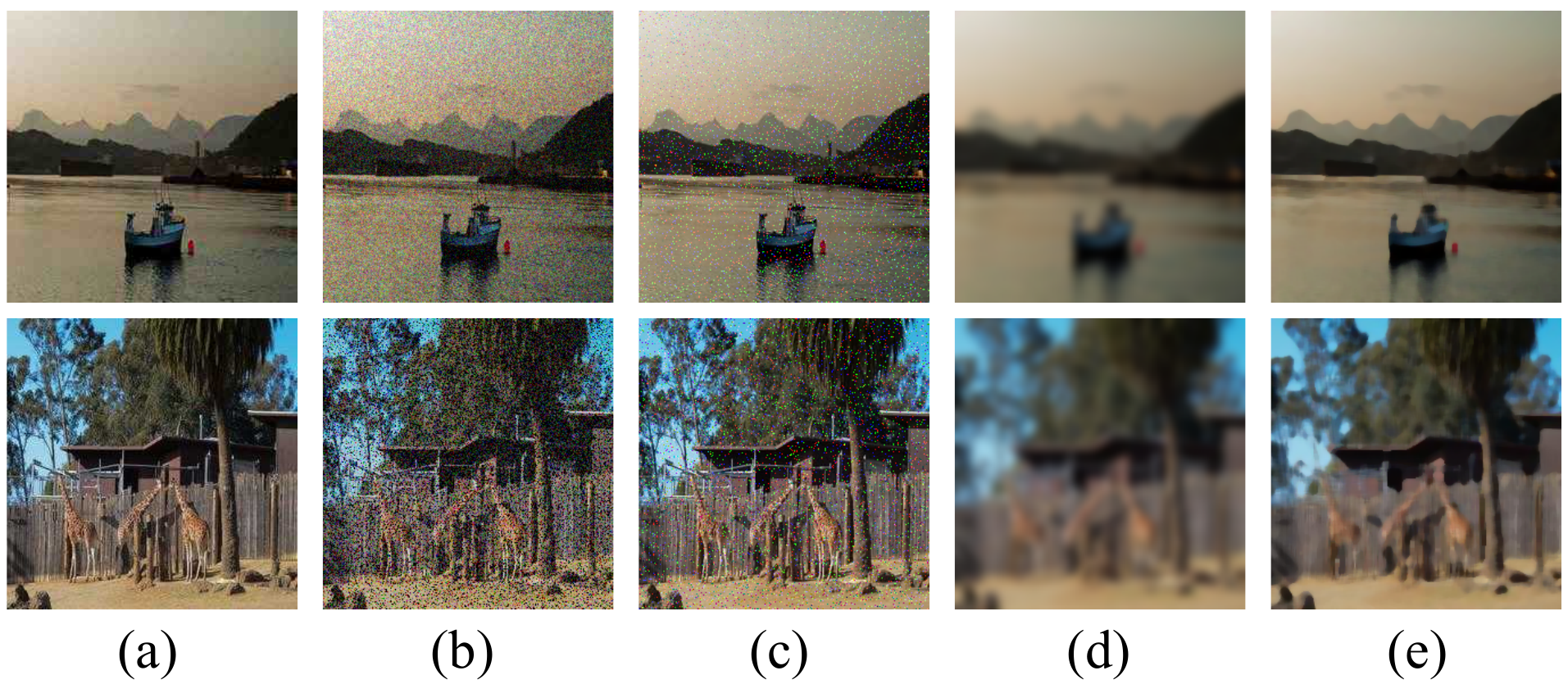}
  \caption{(a) JPEG, $QF$=25; (b) Gaussian noise, $\mu=0, \sigma=0.05$; (c) Salt-and-pepper noise, $p=0.05$; (d) Gaussian blur, $r=4$; (e) Median filter, $k=7$.}
  \label{fig:fig_4}
\end{figure}

\begin{figure}[t!]
  \centering
  \includegraphics[width=\linewidth]{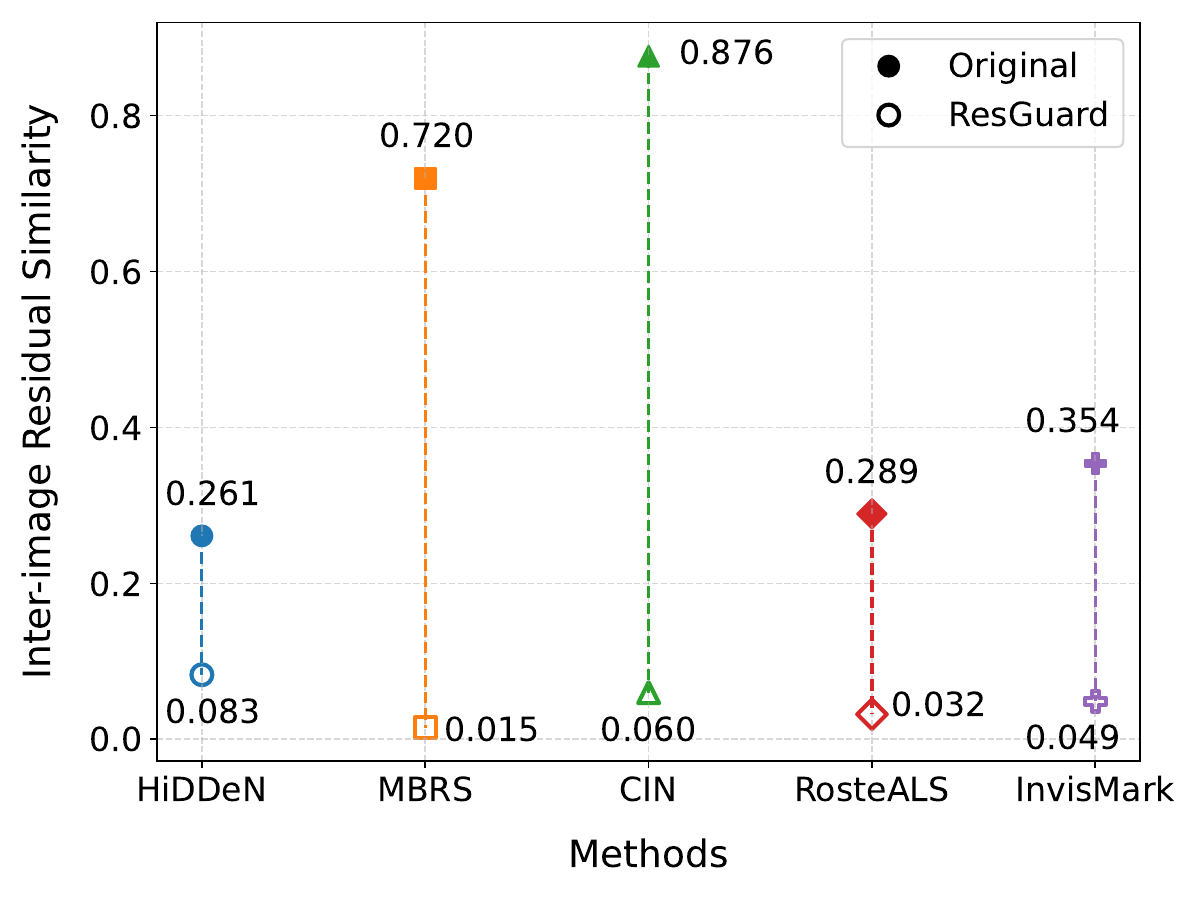}
  \caption{Comparison of inter-image residual similarity before and after applying ResGuard across five watermarking methods. 
  Solid markers denote original models, while hollow markers represent ResGuard-enhanced counterparts. 
  A lower similarity indicates more image-specific residuals.}
  \label{fig:fig_5}
\end{figure}

\begin{figure}[t!]
  \centering
  \includegraphics[width=\linewidth]{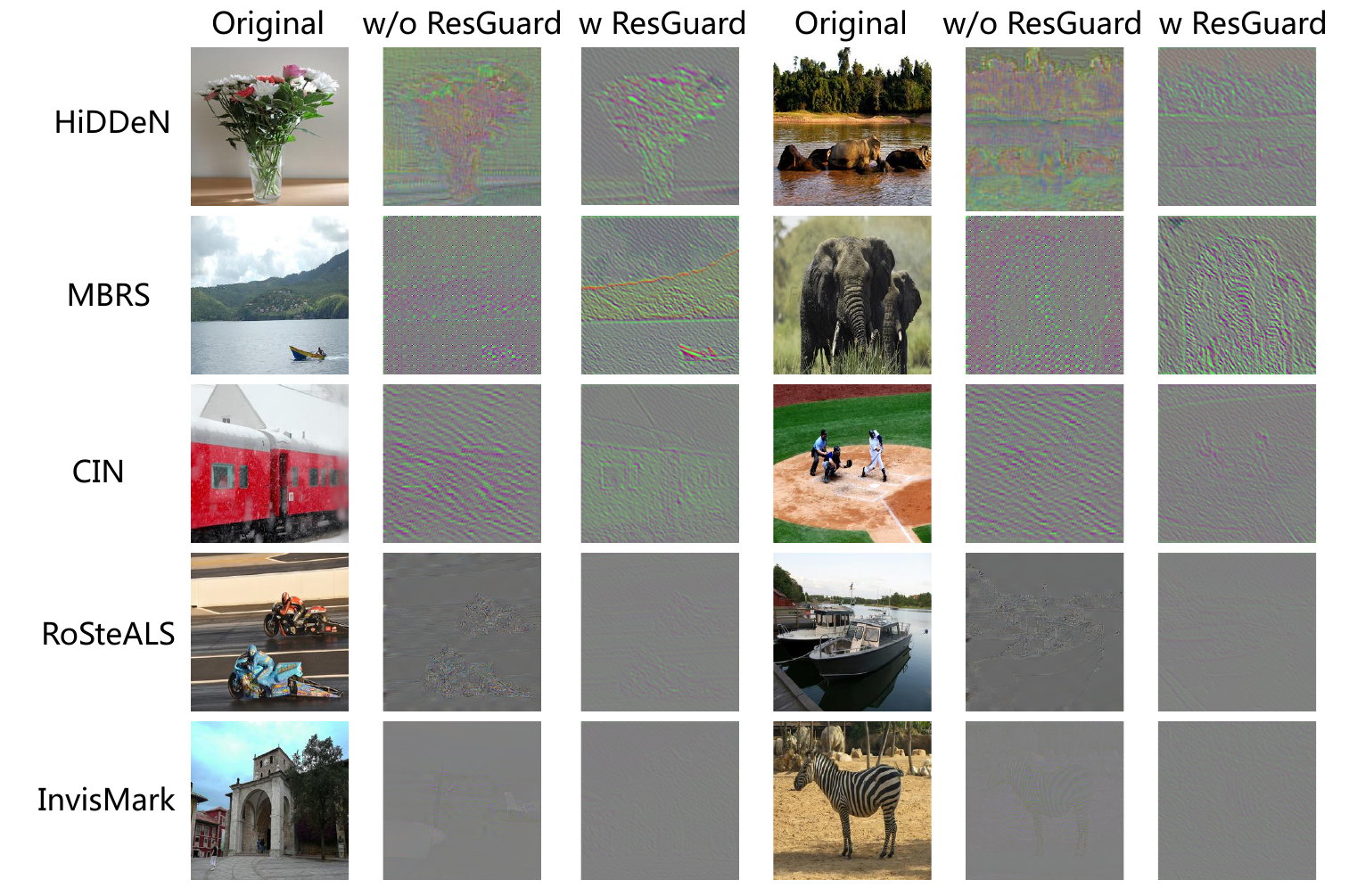}
  \caption{Qualitative comparison of residual patterns before and after applying ResGuard across five watermarking methods. 
  For each method, we visualize the original image, the residual produced by the baseline model (w/o ResGuard), and the residual produced by the ResGuard-enhanced model (w/ ResGuard). 
  All residuals are scaled by a factor of 10 for visualization.}
  \label{fig:fig_6}
\end{figure}

\subsection{KOA Robustness under Diverse Distortions}

While the above experiment focuses on Gaussian noise to isolate the effects of random perturbations, real-world images often suffer from a variety of distortions such as compression and blurring. We therefore further examine whether our method can enhance KOA robustness while maintaining performance under these distortions. To this end, we consider five representative distortion types for a comprehensive evaluation, as shwon in Fig.~\ref{fig:fig_4} We use MBRS as a representative baseline and retrain it under the extended distortion configuration, both with and without our proposed enhancements. The results are summarized in Table~\ref{tab:tab_3}.

Experimental results show that our method significantly improves KOA robustness by approximately 42.69\%, achieving a watermark extraction accuracy of 100.0\% under KOA, while fully preserving the original extraction performance of MBRS under common image degradations, where the average extraction accuracy remains above 99.7\% across all distortions. This demonstrates that our approach not only strengthens defense against KOA but also maintains resilience to typical channel distortions.

\subsection{Evaluations of Residual Image-Specificity}

Beyond robustness evaluation, we further investigate the underlying reason why ResGuard improves KOA resistance by analyzing the image-specificity of embedding residuals. We quantitatively compute the pairwise cosine similarity between residuals generated by embedding the same watermark message into different host images. A lower similarity indicates that the residuals are more uniquely tailored to the specific content of each image. 

As illustrated in Fig.~\ref{fig:fig_5}, our ResGuard-enhanced models reduce inter-image residual similarity across all baselines, with an average decrease of 45.22\%. Specifically, we observe relative reductions of 0.178, 0.705, 0.816, 0.257, and 0.305 over HiDDeN, MBRS, CIN, RoSteALS, and InvisMark, respectively. This demonstrates that the residuals become more closely aligned with the distinct characteristics of each host image, thereby improving image-specificity and playing a critical role in bolstering robustness against KOA.

Furthermore, as shown in Fig.~\ref{fig:fig_6}, the residuals generated by ResGuard-enhanced models exhibit more content-adaptive and diverse structures compared to the original models. These results confirm that our method effectively promotes highly image-specific embeddings, preventing residuals from generalizing across different host images and thereby enhancing robustness against KOA.

\begin{figure*}[t!]
  \centering
  \includegraphics[width=\textwidth]{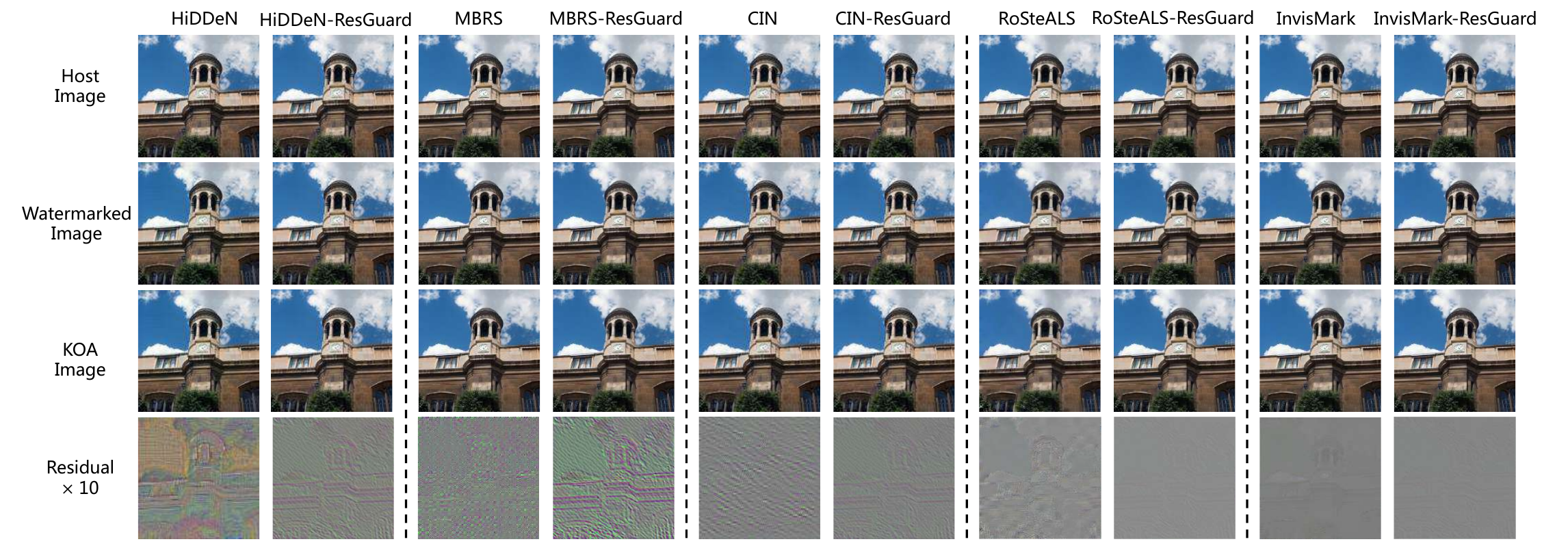}
  \caption{Visual results of original and ResGuard-enhanced models. 
  For each baseline, we show the host image (top), watermarked image, KOA-attacked image, and the residual (bottom, ×10 magnified). 
  Each column pair contrasts the original model (left) and its ResGuard-enhanced variant (right).}
  \label{fig:fig_7}
\end{figure*}

\begin{table}[t]
\centering
\resizebox{\linewidth}{!}{
\begin{tabular}{l|ccc}
\toprule
 
 Methods & SSIM $\Uparrow$ & PSNR $\Uparrow$ & LPIPS $\Downarrow$ \\ 
 
 \midrule
 
 HiDDeN & 36.40 & 0.9474 & 0.0048  \\
 
 \textbf{HiDDeN-ResGuard} & 36.42 & 0.9412 & 0.0045 \\    
 
 \midrule
 
 MBRS & 40.42 & 0.9466 & 0.0123  \\
 
 \textbf{MBRS-ResGuard} & 40.46 & 0.9438 & 0.0115 \\
 
 \midrule

 CIN & 41.29 & 0.9615 & 0.0008  \\
 
 \textbf{CIN-ResGuard} & 41.39 & 0.9587 & 0.0007 \\
 
 \midrule
 
 RosteALS & 34.91 & 0.8970 & 0.0102 \\

 \textbf{RosteALS-ResGuard} & 34.72 & 0.8861 & 0.0103 \\
 
 \midrule
 
 InvisMark & 45.62 & 0.9917 & 0.0005 \\ 
 \textbf{InvisMark-ResGuard} & 45.87 & 0.9913 & 0.0005 \\

\bottomrule
\end{tabular}
}
\vspace{6pt}
\caption{Visual quality of watermarked images. PSNR, SSIM, and LPIPS are reported for original and ResGuard-enhanced models.}
\label{tab:tab_4}
\end{table}

\begin{table}[t]
\centering
\resizebox{\linewidth}{!}{
\begin{tabular}{l|ccc}
\toprule
 
 Methods & SSIM $\Uparrow$ & PSNR $\Uparrow$ & LPIPS $\Downarrow$ \\ 
 
 \midrule
 
 HiDDeN & 36.85 & 0.9331 & 0.0065  \\
 
 \textbf{HiDDeN-ResGuard} & 36.35 & 0.9265 & 0.0062 \\    
 
 \midrule
 
 MBRS & 40.43 & 0.9369  & 0.0167 \\
 
 \textbf{MBRS-ResGuard} & 40.45 & 0.9393 & 0.0171 \\
 
 \midrule

 CIN & 41.33 & 0.9526 & 0.0009  \\
 
 \textbf{CIN-ResGuard} & 41.40 & 0.9532 & 0.0009  \\
 
 \midrule
 
 RosteALS & 34.34 & 0.8891 & 0.0187\\

 \textbf{RosteALS-ResGuard} & 34.42 & 0.8815 & 0.0176\\
 
 \midrule
 
 InvisMark & 45.87 & 0.9813 & 0.0016 \\ 
 \textbf{InvisMark-ResGuard} & 45.85 & 0.9876 & 0.0015 \\

\bottomrule
\end{tabular}
}
\vspace{6pt}
\caption{Visual quality of images after KOA. PSNR, SSIM, and LPIPS are reported for original and ResGuard-enhanced models.}
\label{tab:tab_5}
\end{table}

\begin{table*}[h]
\centering
\resizebox{0.8\linewidth}{!}{
\begin{tabular}{l|c|ccc|cc|c}
\toprule
\multirow{3}{*}{Methods} & \multirow{3}{*}{Mechanisms} &\multicolumn{5}{c}{Watermarking Performance} & \multicolumn{1}{c}{KOA Robustness} \\

\cmidrule{3-7} \cmidrule{8-8}

& & \multicolumn{3}{c|}{Image Quality} & \multicolumn{2}{c|}{Extraction Accuracy} & \multirow{2}{*}{Bit Acc. $\Uparrow$}\\ 
\cmidrule{3-7}
& & PSNR $\Uparrow$ & SSIM $\Uparrow$ & \multicolumn{1}{c|}{LPIPS $\Downarrow$} & Cln. $\Uparrow$ & Dis. $\Uparrow$ \\
\midrule

\multirow{4}{*}{HiDDeN} & \textit{Base} & 36.4045 & 0.9474 & 0.0048 & 1.0000 & 1.0000 & 0.5217 \\ 
& \textit{RSE} & 36.8094 & 0.9102 & 0.0039 & 1.0000 & 1.0000 & 0.9513\\
& \textit{KNL} & 36.3545 & 0.9121 & 0.0050 & 1.0000 & 1.0000 & 0.9601\\
& \textit{ResGuard} & 36.4233 & 0.9472 & 0.0045 & 1.0000 & 1.0000 & 0.9957 \\

\midrule

\multirow{4}{*}{MBRS} & \textit{Base} & 40.4153 & 0.9466 & 0.0123 & 1.0000 & 0.9967 & 0.5927 \\
& \textit{RSE} & 40.2240 & 0.9540 & 0.0109 & 1.0000 & 1.0000 & 0.9633 \\
& \textit{KNL} & 40.0939 & 0.9453 & 0.0117 & 1.0000 & 1.0000 & 0.9589 \\
& \textit{ResGuard} & 40.4627 & 0.9438 & 0.0115 & 1.0000 & 1.0000 & 0.9988 \\

\midrule

\multirow{4}{*}{CIN} & \textit{Base} & 41.2858 & 0.9615 & 0.0008 & 1.0000 & 0.9997 & 0.6147 \\
& \textit{RSE} & 41.3161 & 0.9614 & 0.0008 & 1.0000 & 0.9997 & 0.9782 \\
& \textit{KNL} & 41.1667 & 0.9626 & 0.0008 & 1.0000 & 0.9996 & 0.9694 \\
& \textit{ResGuard} & 41.3865 & 0.9587 & 0.0007 & 1.0000 & 0.9997 & 0.9983 \\

\midrule

\multirow{4}{*}{RosteALS} & \textit{Base} & 34.9123 & 0.8970 & 0.0102 & 1.0000 & 0.9926 & 0.6332 \\
& \textit{RSE} & 34.8531 & 0.9002 & 0.0100 & 1.0000 & 0.9929 & 0.9517 \\
& \textit{KNL} & 34.6826 & 0.8872 & 0.0116 & 1.0000 & 0.9927 & 0.9648 \\
& \textit{ResGuard} & 34.7246 & 0.8861 & 0.0103 & 1.0000 & 0.9931 & 0.9987 \\

\midrule

\multirow{4}{*}{InvisMark} & \textit{Base} & 45.6226 & 0.9917 & 0.0005 & 1.0000 & 1.0000 & 0.6312 \\
& \textit{RSE} & 45.9489 & 0.9936 & 0.0004 & 1.0000 & 1.0000 & 0.9765 \\
& \textit{KNL} & 45.2108 & 0.9902 & 0.0006 & 1.0000 & 1.0000 & 0.9639 \\
& \textit{ResGuard} & 45.8715 & 0.9913 & 0.0005 & 1.0000 & 1.0000 & 0.9992 \\

\bottomrule
\end{tabular}
}
\caption{Ablation study on the effectiveness of the residual specificity enhancement loss and the residual transferability suppression loss. “Cln” indicates results on clean images, while “Dis” represents results under channel distortions.}
\label{tab:tab_6}
\vspace{-10pt}
\end{table*}

\subsection{Visual Quality and Imperceptibility}

\paragraph{Watermarking Imperceptibility.}
We first verify that our image-specific residual watermarking strategy, designed to enhance robustness against KOA, does not compromise the visual quality of watermarked images. As shown in Table~\ref{tab:tab_4}, we report PSNR, SSIM, and LPIPS for both the original and ResGuard-enhanced models. Across all baselines, the ResGuard-integrated variants achieve perceptual quality comparable to their original counterparts. On average, the absolute change introduced by ResGuard is negligible: PSNR varies by less than 0.1 dB, SSIM differs by under 0.003, and LPIPS changes by no more than 0.0006. 
Qualitative comparisons in Fig.~\ref{fig:fig_7} further confirm that the watermarked images remain visually indistinguishable from the host images. 
While the residual maps differ structurally due to the image-specific embedding strategy, they remain visually sparse and low in magnitude. These results demonstrate that integrating ResGuard does not degrade the visual quality of the watermarked images.

\paragraph{Perceptual Quality under KOA.}
We further assess the perceptual impact of KOA by evaluating the visual quality of watermarked images after the attack. 
As shown in Table~\ref{tab:tab_5}, the attacked images for all methods, both in their original form and with ResGuard integration, exhibit negligible changes in visual quality. The reported differences in PSNR, SSIM, and LPIPS are minimal, typically within 0.01 dB, 0.002, and 0.003, respectively. These results confirm that KOA does not introduce visible artifacts and that the attacked images remain visually indistinguishable from the watermarked versions. The imperceptibility of this attack underscores its deceptive nature and further emphasizes the importance of improving watermark robustness through content-dependent embedding rather than relying on visual cues.

\subsection{Ablation Study}

\paragraph{Effectiveness of RSE and KOA Noise Layer.} 
In ResGuard, we introduce two key components to explicitly promote highly image-specific embedding patterns, thereby improving robustness against KOA. 
To evaluate their individual contributions, we retrain all baseline watermarking methods under four configurations: (1) without RSE and KOA noise layer (“Base”), (2) with only RSE (“RSE”), (3) with only KOA noise layer (“KNL”), and (4) with both (“ResGuard”). 
The results are summarized in Table~\ref{tab:tab_6}.

Applying either RSE or KNL individually leads to substantial improvements in KOA bit accuracy compared to the baseline. Across HiDDeN, MBRS, CIN, RoSteALS, and InvisMark, employing RSE increases the average KOA bit accuracy by 42.96\%, 37.06\%, 36.35\%, 31.85\%, and 34.53\%, respectively, while KNL yields comparable gains of 43.84\%, 36.62\%, 35.47\%, 33.16\%, and 33.27\%. RSE enhances the dependence of embedding residuals on host content through contrastive regularization, producing more image-specific patterns, whereas KNL reduces cross-image residual transferability by simulating and defending against residual-based perturbations during training. 
When combined, both modules deliver complementary benefits, achieving average improvements of 36.51\% across methods and nearly 100\% KOA bit accuracy overall.  Importantly, ResGuard maintains visual quality and performance under common distortions, demonstrating an excellent balance between robustness and imperceptibility.

\begin{figure*}[htbp]
    \centering
    \includegraphics[width=0.19\textwidth]{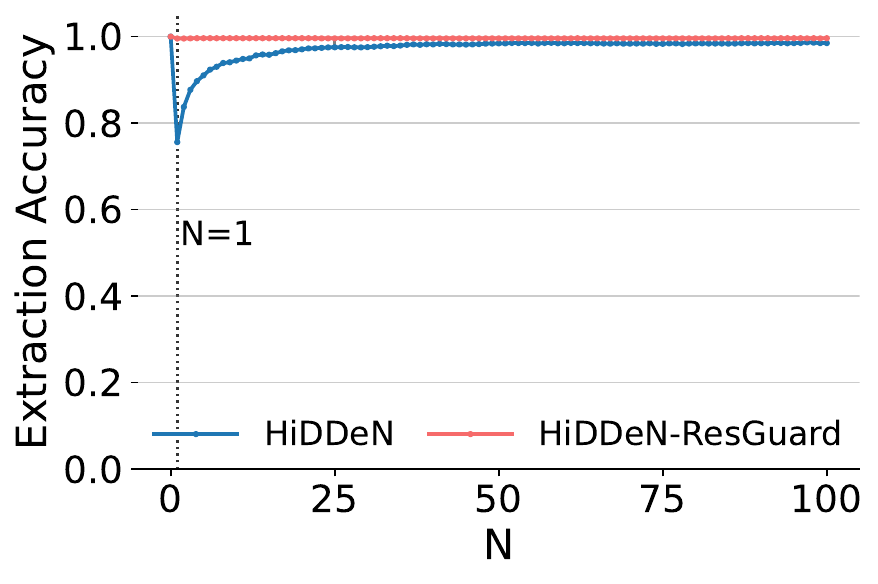}
    \includegraphics[width=0.19\textwidth]{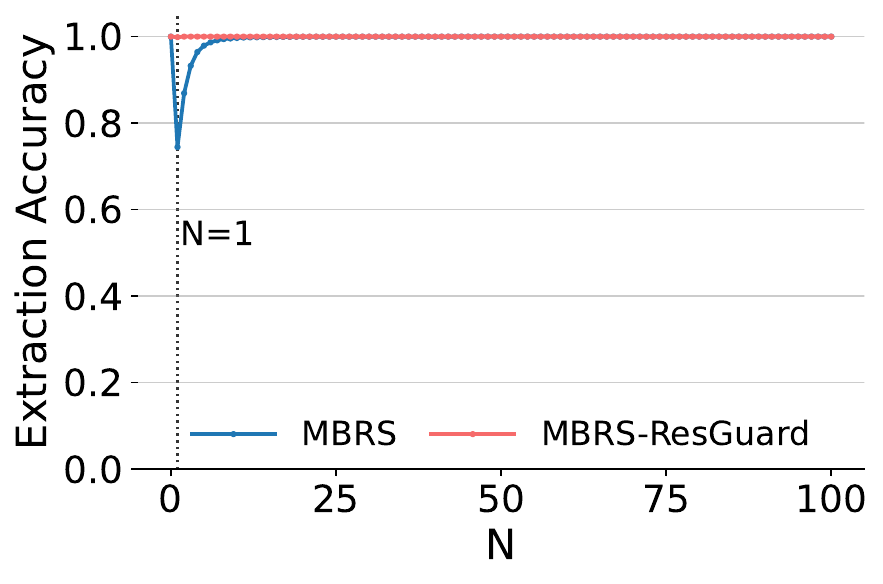}
    \includegraphics[width=0.19\textwidth]{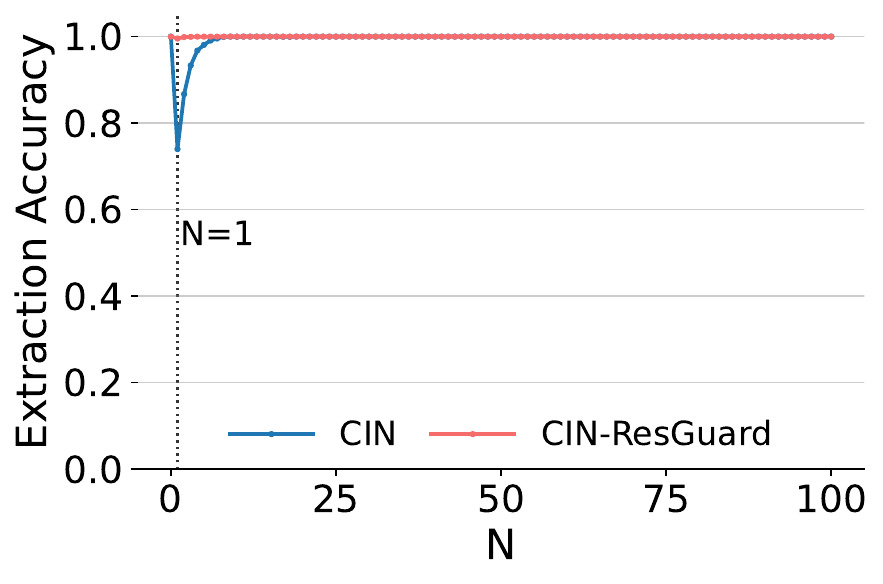}
    \includegraphics[width=0.19\textwidth]{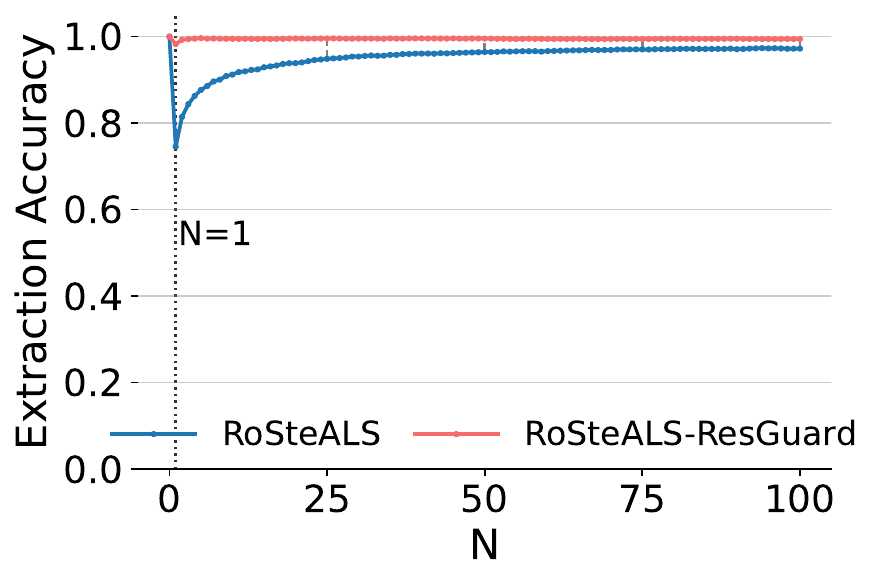}
    \includegraphics[width=0.19\textwidth]{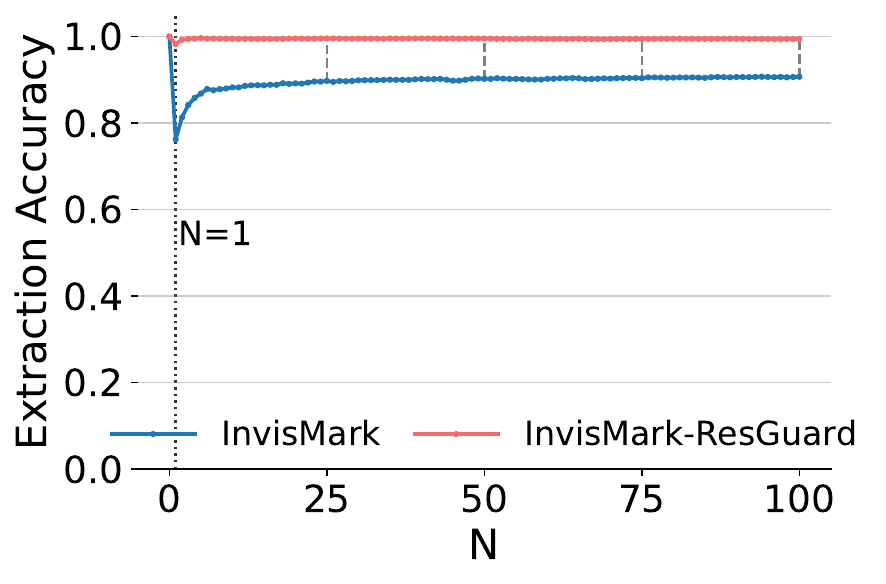}

    \caption{Bitwise extraction accuracy under KOA across five baseline methods. Each curve shows performance variation with the number of available host--watermarked pairs \(N\). Models equipped with ResGuard maintain consistently high accuracy, demonstrating strong robustness against KOA.}
    \label{fig:fig_8}
\end{figure*}

\begin{table}[t]
\centering
\resizebox{\linewidth}{!}{
\begin{tabular}{l|cc|c}
\toprule
 \multirow{2}{*}{Methods} & \multicolumn{2}{c|}{Extraction Accuracy} & \multicolumn{1}{c}{KOA Robustness} \\
 
\cmidrule(lr){2-4} 
 
 & Cln. $\Uparrow$ & Dis. $\Uparrow$ & Bit Acc. $\Uparrow$ \\ 
 
 \midrule
 
 HiDDeN & 1.0000 & 1.0000 & 0.7333 \\
 
 \textbf{HiDDeN-ResGuard} & 1.0000 & 1.0000 & \textbf{1.0000} \\    
 
 \midrule
 
 MBRS & 1.0000 & 1.0000 & 0.7447 \\
 
 \textbf{MBRS-ResGuard} & 1.0000 & 1.0000 & \textbf{1.0000} \\
 
 \midrule
 
 CIN & 1.0000 & 0.9997 & 0.7313 \\
 
 \textbf{CIN-ResGuard} & 1.0000 & 0.9997 & \textbf{0.9983} \\
 
 \midrule
 
 RosteALS & 1.0000 & 0.9926 & 0.7180 \\

 \textbf{RosteALS-ResGuard} & 1.0000 & 0.9931 & \textbf{0.9987} \\
 
 \midrule
 
 InvisMark & 1.0000 & 1.0000 & 0.7673 \\ 
 
 \textbf{InvisMark-ResGuard} & 1.0000 & 1.0000 & \textbf{1.0000} \\

\bottomrule
\end{tabular}
}
\vspace{6pt}
\caption{Comparison results of KOA robustness. “Cln” indicates results on clean images, while “Dis” represents results under channel distortions.}
\label{tab:tab_7}
\end{table}

\subsection{Effect of Watermark Message Variability}

Beyond quantitative improvements, we further analyze how watermark message variability affects KOA effectiveness. In our experimental setting, all host--watermarked pairs embed the same message, reflecting the common practice of reusing a single watermark key for ownership verification. To further examine whether KOA remains effective when messages vary across images, we additionally evaluate a different-message setting, where each host image is embedded with an independent watermark message.

As shown in Fig.~\ref{fig:fig_8}, KOA remains effective even with only one known host--watermarked pair ($N=1$), the average bit accuracy of baseline methods drops from nearly 100\% to around 70\%. This indicates that even minimal knowledge of a single pair allows the adversary to substantially degrade watermark extraction accuracy. However, unlike the same-message setting, the attack becomes weaker as $N$ increases. The reason is that message-dependent residual components gradually cancel out during averaging, so the estimated residual approaches zero-mean random noise, to which existing watermarking models are relatively robust.

These results suggest that the main vulnerability of KOA does not come from the perturbation magnitude itself, but from the cross-image similarity of residuals, especially when identical watermark messages are embedded. In both same-message and different-message settings, ResGuard consistently maintains high extraction accuracy by enforcing image-specific embedding patterns. 

We further report detailed results under the minimal-case setting of $N=1$, where the attacker has access to only a single host--watermarked pair. This represents a practical and challenging scenario, requiring minimal prior knowledge while still causing significant degradation. Because KOA essentially subtracts an additive residual, which behaves like a quasi-random perturbation, we use Gaussian noise in the standard noise layer to ensure consistency across models. The results in Table~\ref{tab:tab_7} show that ResGuard consistently improves robustness across all baselines, boosting bitwise extraction accuracy from about 70\% to nearly 100\%. These results demonstrate that ResGuard effectively mitigates KOA while preserving watermark fidelity and imperceptibility.


\section{Conclusion}
This paper reveals a critical yet overlooked vulnerability in deep learning-based image watermarking: their susceptibility to the Known Original Attack (KOA), where an adversary can effectively remove the embedded watermark using only one or a few original–watermarked image pairs. We attribute this vulnerability to the insufficient image specificity of embedding residuals generated by END-based frameworks. To address this, we propose ResGuard, a plug-and-play module that enforces image-dependent embedding through two complementary components: a residual specificity enhancement loss that strengthens image-specific residuals, and a KOA noise layer that simulates residual-based attacks during training. Extensive experiments show that ResGuard substantially improves robustness against KOA while fully preserving visual imperceptibility and extraction accuracy.

\bibliographystyle{ACM-Reference-Format}
\bibliography{main}


\begin{thebibliography}{26}


\ifx \showCODEN    \undefined \def \showCODEN     #1{\unskip}     \fi
\ifx \showISBNx    \undefined \def \showISBNx     #1{\unskip}     \fi
\ifx \showISBNxiii \undefined \def \showISBNxiii  #1{\unskip}     \fi
\ifx \showISSN     \undefined \def \showISSN      #1{\unskip}     \fi
\ifx \showLCCN     \undefined \def \showLCCN      #1{\unskip}     \fi
\ifx \shownote     \undefined \def \shownote      #1{#1}          \fi
\ifx \showarticletitle \undefined \def \showarticletitle #1{#1}   \fi
\ifx \showURL      \undefined \def \showURL       {\relax}        \fi
\providecommand\bibfield[2]{#2}
\providecommand\bibinfo[2]{#2}
\providecommand\natexlab[1]{#1}
\providecommand\showeprint[2][]{arXiv:#2}

\bibitem[Agustsson and Timofte(2017)]%
        {agustsson2017ntire}
\bibfield{author}{\bibinfo{person}{Eirikur Agustsson} {and}
  \bibinfo{person}{Radu Timofte}.} \bibinfo{year}{2017}\natexlab{}.
\newblock \showarticletitle{Ntire 2017 challenge on single image
  super-resolution: Dataset and study}. In
  \bibinfo{booktitle}{\emph{Proceedings of the IEEE conference on computer
  vision and pattern recognition workshops}}. \bibinfo{pages}{126--135}.
\newblock


\bibitem[Alotaibi and Elrefaei(2019)]%
        {alotaibi2019text}
\bibfield{author}{\bibinfo{person}{Reem~A Alotaibi} {and}
  \bibinfo{person}{Lamiaa~A Elrefaei}.} \bibinfo{year}{2019}\natexlab{}.
\newblock \showarticletitle{Text-image watermarking based on integer wavelet
  transform (IWT) and discrete cosine transform (DCT)}.
\newblock \bibinfo{journal}{\emph{Applied Computing and Informatics}}
  \bibinfo{volume}{15}, \bibinfo{number}{2} (\bibinfo{year}{2019}),
  \bibinfo{pages}{191--202}.
\newblock


\bibitem[Bui et~al\mbox{.}(2023a)]%
        {bui2023trustmark}
\bibfield{author}{\bibinfo{person}{Tu Bui}, \bibinfo{person}{Shruti Agarwal},
  {and} \bibinfo{person}{John Collomosse}.} \bibinfo{year}{2023}\natexlab{a}.
\newblock \showarticletitle{Trustmark: Universal watermarking for arbitrary
  resolution images}.
\newblock \bibinfo{journal}{\emph{arXiv preprint arXiv:2311.18297}}
  (\bibinfo{year}{2023}).
\newblock


\bibitem[Bui et~al\mbox{.}(2023b)]%
        {bui2023rosteals}
\bibfield{author}{\bibinfo{person}{Tu Bui}, \bibinfo{person}{Shruti Agarwal},
  \bibinfo{person}{Ning Yu}, {and} \bibinfo{person}{John Collomosse}.}
  \bibinfo{year}{2023}\natexlab{b}.
\newblock \showarticletitle{Rosteals: Robust steganography using autoencoder
  latent space}. In \bibinfo{booktitle}{\emph{Proceedings of the IEEE/CVF
  conference on computer vision and pattern recognition}}.
  \bibinfo{pages}{933--942}.
\newblock


\bibitem[Cayre et~al\mbox{.}(2004)]%
        {cayre2004watermarking}
\bibfield{author}{\bibinfo{person}{Fran{\c{c}}ois Cayre},
  \bibinfo{person}{Caroline Fontaine}, {and} \bibinfo{person}{Teddy Furon}.}
  \bibinfo{year}{2004}\natexlab{}.
\newblock \showarticletitle{Watermarking attack: Security of wss techniques}.
  In \bibinfo{booktitle}{\emph{International Workshop on Digital
  Watermarking}}. Springer, \bibinfo{pages}{171--183}.
\newblock


\bibitem[Cox et~al\mbox{.}(2007)]%
        {cox2007digital}
\bibfield{author}{\bibinfo{person}{Ingemar Cox}, \bibinfo{person}{Matthew
  Miller}, \bibinfo{person}{Jeffrey Bloom}, \bibinfo{person}{Jessica Fridrich},
  {and} \bibinfo{person}{Ton Kalker}.} \bibinfo{year}{2007}\natexlab{}.
\newblock \bibinfo{booktitle}{\emph{Digital watermarking and steganography}}.
\newblock \bibinfo{publisher}{Morgan kaufmann}.
\newblock


\bibitem[Esser et~al\mbox{.}(2021)]%
        {esser2021taming}
\bibfield{author}{\bibinfo{person}{Patrick Esser}, \bibinfo{person}{Robin
  Rombach}, {and} \bibinfo{person}{Bjorn Ommer}.}
  \bibinfo{year}{2021}\natexlab{}.
\newblock \showarticletitle{Taming transformers for high-resolution image
  synthesis}. In \bibinfo{booktitle}{\emph{Proceedings of the IEEE/CVF
  conference on computer vision and pattern recognition}}.
  \bibinfo{pages}{12873--12883}.
\newblock


\bibitem[Fang et~al\mbox{.}(2022)]%
        {fang2022pimog}
\bibfield{author}{\bibinfo{person}{Han Fang}, \bibinfo{person}{Zhaoyang Jia},
  \bibinfo{person}{Zehua Ma}, \bibinfo{person}{Ee-Chien Chang}, {and}
  \bibinfo{person}{Weiming Zhang}.} \bibinfo{year}{2022}\natexlab{}.
\newblock \showarticletitle{PIMoG: An effective screen-shooting noise-layer
  simulation for deep-learning-based watermarking network}. In
  \bibinfo{booktitle}{\emph{Proceedings of the 30th ACM international
  conference on multimedia}}. \bibinfo{pages}{2267--2275}.
\newblock


\bibitem[Fang et~al\mbox{.}(2023)]%
        {fang2023flow}
\bibfield{author}{\bibinfo{person}{Han Fang}, \bibinfo{person}{Yupeng Qiu},
  \bibinfo{person}{Kejiang Chen}, \bibinfo{person}{Jiyi Zhang},
  \bibinfo{person}{Weiming Zhang}, {and} \bibinfo{person}{Ee-Chien Chang}.}
  \bibinfo{year}{2023}\natexlab{}.
\newblock \showarticletitle{Flow-based robust watermarking with invertible
  noise layer for black-box distortions}. In
  \bibinfo{booktitle}{\emph{Proceedings of the AAAI conference on artificial
  intelligence}}, Vol.~\bibinfo{volume}{37}. \bibinfo{pages}{5054--5061}.
\newblock


\bibitem[Fang et~al\mbox{.}(2018)]%
        {fang2018screen}
\bibfield{author}{\bibinfo{person}{Han Fang}, \bibinfo{person}{Weiming Zhang},
  \bibinfo{person}{Hang Zhou}, \bibinfo{person}{Hao Cui}, {and}
  \bibinfo{person}{Nenghai Yu}.} \bibinfo{year}{2018}\natexlab{}.
\newblock \showarticletitle{Screen-shooting resilient watermarking}.
\newblock \bibinfo{journal}{\emph{IEEE Transactions on Information Forensics
  and Security}} \bibinfo{volume}{14}, \bibinfo{number}{6}
  (\bibinfo{year}{2018}), \bibinfo{pages}{1403--1418}.
\newblock


\bibitem[Fernandez et~al\mbox{.}(2022)]%
        {fernandez2022watermarking}
\bibfield{author}{\bibinfo{person}{Pierre Fernandez},
  \bibinfo{person}{Alexandre Sablayrolles}, \bibinfo{person}{Teddy Furon},
  \bibinfo{person}{Herv{\'e} J{\'e}gou}, {and} \bibinfo{person}{Matthijs
  Douze}.} \bibinfo{year}{2022}\natexlab{}.
\newblock \showarticletitle{Watermarking images in self-supervised latent
  spaces}. In \bibinfo{booktitle}{\emph{ICASSP 2022-2022 IEEE International
  Conference on Acoustics, Speech and Signal Processing (ICASSP)}}. IEEE,
  \bibinfo{pages}{3054--3058}.
\newblock


\bibitem[Hamidi et~al\mbox{.}(2018)]%
        {hamidi2018hybrid}
\bibfield{author}{\bibinfo{person}{Mohamed Hamidi}, \bibinfo{person}{Mohamed~El
  Haziti}, \bibinfo{person}{Hocine Cherifi}, {and} \bibinfo{person}{Mohammed~El
  Hassouni}.} \bibinfo{year}{2018}\natexlab{}.
\newblock \showarticletitle{Hybrid blind robust image watermarking technique
  based on DFT-DCT and Arnold transform}.
\newblock \bibinfo{journal}{\emph{Multimedia Tools and Applications}}
  \bibinfo{volume}{77} (\bibinfo{year}{2018}), \bibinfo{pages}{27181--27214}.
\newblock


\bibitem[Hu et~al\mbox{.}(2014)]%
        {hu2014orthogonal}
\bibfield{author}{\bibinfo{person}{Hai-tao Hu}, \bibinfo{person}{Ya-dong
  Zhang}, \bibinfo{person}{Chao Shao}, {and} \bibinfo{person}{Quan Ju}.}
  \bibinfo{year}{2014}\natexlab{}.
\newblock \showarticletitle{Orthogonal moments based on exponent functions:
  Exponent-Fourier moments}.
\newblock \bibinfo{journal}{\emph{Pattern Recognition}} \bibinfo{volume}{47},
  \bibinfo{number}{8} (\bibinfo{year}{2014}), \bibinfo{pages}{2596--2606}.
\newblock


\bibitem[Hu(1962)]%
        {hu1962visual}
\bibfield{author}{\bibinfo{person}{Ming-Kuei Hu}.}
  \bibinfo{year}{1962}\natexlab{}.
\newblock \showarticletitle{Visual pattern recognition by moment invariants}.
\newblock \bibinfo{journal}{\emph{IRE transactions on information theory}}
  \bibinfo{volume}{8}, \bibinfo{number}{2} (\bibinfo{year}{1962}),
  \bibinfo{pages}{179--187}.
\newblock


\bibitem[Jia et~al\mbox{.}(2021)]%
        {jia2021mbrs}
\bibfield{author}{\bibinfo{person}{Zhaoyang Jia}, \bibinfo{person}{Han Fang},
  {and} \bibinfo{person}{Weiming Zhang}.} \bibinfo{year}{2021}\natexlab{}.
\newblock \showarticletitle{Mbrs: Enhancing robustness of dnn-based
  watermarking by mini-batch of real and simulated jpeg compression}. In
  \bibinfo{booktitle}{\emph{Proceedings of the 29th ACM international
  conference on multimedia}}. \bibinfo{pages}{41--49}.
\newblock


\bibitem[Kang et~al\mbox{.}(2003)]%
        {kang2003dwt}
\bibfield{author}{\bibinfo{person}{Xiangui Kang}, \bibinfo{person}{Jiwu Huang},
  \bibinfo{person}{Yun~Q Shi}, {and} \bibinfo{person}{Yan Lin}.}
  \bibinfo{year}{2003}\natexlab{}.
\newblock \showarticletitle{A DWT-DFT composite watermarking scheme robust to
  both affine transform and JPEG compression}.
\newblock \bibinfo{journal}{\emph{IEEE transactions on circuits and systems for
  video technology}} \bibinfo{volume}{13}, \bibinfo{number}{8}
  (\bibinfo{year}{2003}), \bibinfo{pages}{776--786}.
\newblock


\bibitem[Lin et~al\mbox{.}(2014)]%
        {lin2014microsoft}
\bibfield{author}{\bibinfo{person}{Tsung-Yi Lin}, \bibinfo{person}{Michael
  Maire}, \bibinfo{person}{Serge Belongie}, \bibinfo{person}{James Hays},
  \bibinfo{person}{Pietro Perona}, \bibinfo{person}{Deva Ramanan},
  \bibinfo{person}{Piotr Doll{\'a}r}, {and} \bibinfo{person}{C~Lawrence
  Zitnick}.} \bibinfo{year}{2014}\natexlab{}.
\newblock \showarticletitle{Microsoft coco: Common objects in context}. In
  \bibinfo{booktitle}{\emph{Computer vision--ECCV 2014: 13th European
  conference, zurich, Switzerland, September 6-12, 2014, proceedings, part v
  13}}. Springer, \bibinfo{pages}{740--755}.
\newblock


\bibitem[Ma et~al\mbox{.}(2022)]%
        {ma2022towards}
\bibfield{author}{\bibinfo{person}{Rui Ma}, \bibinfo{person}{Mengxi Guo},
  \bibinfo{person}{Yi Hou}, \bibinfo{person}{Fan Yang}, \bibinfo{person}{Yuan
  Li}, \bibinfo{person}{Huizhu Jia}, {and} \bibinfo{person}{Xiaodong Xie}.}
  \bibinfo{year}{2022}\natexlab{}.
\newblock \showarticletitle{Towards blind watermarking: Combining invertible
  and non-invertible mechanisms}. In \bibinfo{booktitle}{\emph{Proceedings of
  the 30th ACM International Conference on Multimedia}}.
  \bibinfo{pages}{1532--1542}.
\newblock


\bibitem[Mehta et~al\mbox{.}(2016)]%
        {mehta2016lwt}
\bibfield{author}{\bibinfo{person}{Rajesh Mehta}, \bibinfo{person}{Navin
  Rajpal}, {and} \bibinfo{person}{Virendra~P Vishwakarma}.}
  \bibinfo{year}{2016}\natexlab{}.
\newblock \showarticletitle{LWT-QR decomposition based robust and efficient
  image watermarking scheme using Lagrangian SVR}.
\newblock \bibinfo{journal}{\emph{Multimedia Tools and Applications}}
  \bibinfo{volume}{75} (\bibinfo{year}{2016}), \bibinfo{pages}{4129--4150}.
\newblock


\bibitem[Pakdaman et~al\mbox{.}(2017)]%
        {pakdaman2017prediction}
\bibfield{author}{\bibinfo{person}{Zahra Pakdaman}, \bibinfo{person}{Saeid
  Saryazdi}, {and} \bibinfo{person}{Hossein Nezamabadi-Pour}.}
  \bibinfo{year}{2017}\natexlab{}.
\newblock \showarticletitle{A prediction based reversible image watermarking in
  Hadamard domain}.
\newblock \bibinfo{journal}{\emph{Multimedia Tools and Applications}}
  \bibinfo{volume}{76} (\bibinfo{year}{2017}), \bibinfo{pages}{8517--8545}.
\newblock


\bibitem[Soualmi et~al\mbox{.}(2018)]%
        {soualmi2018schur}
\bibfield{author}{\bibinfo{person}{Abdallah Soualmi}, \bibinfo{person}{Adel
  Alti}, {and} \bibinfo{person}{Lamri Laouamer}.}
  \bibinfo{year}{2018}\natexlab{}.
\newblock \showarticletitle{Schur and DCT decomposition based medical images
  watermarking}. In \bibinfo{booktitle}{\emph{2018 Sixth International
  Conference on Enterprise Systems (ES)}}. IEEE, \bibinfo{pages}{204--210}.
\newblock


\bibitem[Su et~al\mbox{.}(2014)]%
        {su2014blind}
\bibfield{author}{\bibinfo{person}{Qingtang Su}, \bibinfo{person}{Yugang Niu},
  \bibinfo{person}{Hailin Zou}, \bibinfo{person}{Yongsheng Zhao}, {and}
  \bibinfo{person}{Tao Yao}.} \bibinfo{year}{2014}\natexlab{}.
\newblock \showarticletitle{A blind double color image watermarking algorithm
  based on QR decomposition}.
\newblock \bibinfo{journal}{\emph{Multimedia tools and applications}}
  \bibinfo{volume}{72} (\bibinfo{year}{2014}), \bibinfo{pages}{987--1009}.
\newblock


\bibitem[Van~Schyndel et~al\mbox{.}(1994)]%
        {van1994digital}
\bibfield{author}{\bibinfo{person}{Ron~G Van~Schyndel},
  \bibinfo{person}{Andrew~Z Tirkel}, {and} \bibinfo{person}{Charles~F
  Osborne}.} \bibinfo{year}{1994}\natexlab{}.
\newblock \showarticletitle{A digital watermark}. In
  \bibinfo{booktitle}{\emph{Proceedings of 1st international conference on
  image processing}}, Vol.~\bibinfo{volume}{2}. IEEE, \bibinfo{pages}{86--90}.
\newblock


\bibitem[Xu et~al\mbox{.}(2025)]%
        {xu2025invismark}
\bibfield{author}{\bibinfo{person}{Rui Xu}, \bibinfo{person}{Mengya Hu},
  \bibinfo{person}{Deren Lei}, \bibinfo{person}{Yaxi Li},
  \bibinfo{person}{David Lowe}, \bibinfo{person}{Alex Gorevski},
  \bibinfo{person}{Mingyu Wang}, \bibinfo{person}{Emily Ching}, {and}
  \bibinfo{person}{Alex Deng}.} \bibinfo{year}{2025}\natexlab{}.
\newblock \showarticletitle{InvisMark: Invisible and Robust Watermarking for
  AI-generated Image Provenance}. In \bibinfo{booktitle}{\emph{2025 IEEE/CVF
  Winter Conference on Applications of Computer Vision (WACV)}}. IEEE,
  \bibinfo{pages}{909--918}.
\newblock


\bibitem[Zhang et~al\mbox{.}(2019)]%
        {zhang2019robust}
\bibfield{author}{\bibinfo{person}{Kevin~Alex Zhang}, \bibinfo{person}{Lei Xu},
  \bibinfo{person}{Alfredo Cuesta-Infante}, {and} \bibinfo{person}{Kalyan
  Veeramachaneni}.} \bibinfo{year}{2019}\natexlab{}.
\newblock \showarticletitle{Robust invisible video watermarking with
  attention}.
\newblock \bibinfo{journal}{\emph{arXiv preprint arXiv:1909.01285}}
  (\bibinfo{year}{2019}).
\newblock


\bibitem[Zhu et~al\mbox{.}(2018)]%
        {zhu2018hidden}
\bibfield{author}{\bibinfo{person}{Jiren Zhu}, \bibinfo{person}{Russell
  Kaplan}, \bibinfo{person}{Justin Johnson}, {and} \bibinfo{person}{Li
  Fei-Fei}.} \bibinfo{year}{2018}\natexlab{}.
\newblock \showarticletitle{Hidden: Hiding data with deep networks}. In
  \bibinfo{booktitle}{\emph{Proceedings of the European conference on computer
  vision (ECCV)}}. \bibinfo{pages}{657--672}.
\newblock


\end{thebibliography}


\end{document}